# Joint order assignment and picking station scheduling in KIVA warehouses with multiple stations

Xiying Yang[a]; Guowei Hua[a,*]; Li Zhang[a]; T.C.E Cheng[b]; Tsan-Ming Choi[c]

[a]School of Economics and Management, Beijing Jiaotong University,
Beijing 100044, China

[b]Department of Logistics and Maritime Studies, The Hong Kong Polytechnic University,
Hung Hom, Hong Kong, China

[c]Centre for Supply Chain Research, Management School, University of Liverpool,
the United Kingdom

**Abstract**

We consider the problem of allocating orders to multiple stations and sequencing the interlinked order and rack processing flows in each station in the robot-assisted KIVA warehouse. The various decisions involved in the problem, which are closely associated and must be solved in real time, are often tackled separately for ease of treatment. However, exploiting the synergy between order assignment and picking station scheduling benefits picking efficiency. We develop a comprehensive mathematical model that takes the synergy into consideration to minimize the total number of rack visits. To solve this intractable problem, we develop an efficient algorithm based on simulated annealing and beam search. Computational studies show that our proposed approach outperforms the rule-based greedy policy and the independent picking station scheduling method in terms of solution quality, saving over one-third and one-fifth of rack visits compared with the former and latter, respectively.

*Corresponding author: Guowei Hua

# 1 Introduction

## 1.1 Background and motivation

At 00:26 on 11 November in 2020 (the Double Eleven Shopping Festival in China), the peak of order creation in T-mall reached 583,000 orders per second (Sina, 2020). The rapid development of e-commerce has brought new challenges to warehouse operations. Order picking plays a crucial role among all these operations, which directly affects the overall order fulfillment efficiency (Lamballais et al., 2017; Shen et al., 2020). The Robotic Mobile Fulfillment System (RMFS) is invented to improve order picking efficiency and reduce labour costs by exploiting rack-moving mobile robots (Boysen et al., 2017). The cooperation between the robots and movable racks eliminates pickers' unproductive movement in the picker-to-parts system (Battini et al., 2017). Compared with traditional manual warehouses, the picking performance of RMFS is far superior, which is reported to achieve over 600 order-lines per hour per workstation (Wulfraat, 2012; Banker, 2016). Nevertheless, order picking in RMFS needs further efficiency improvement due to the growing demand and increasingly tight delivery schedules brought by the prosperity of e-commerce (Batt & Gallino, 2019; Azadeh et al., 2017; Zhuang et al., 2021).

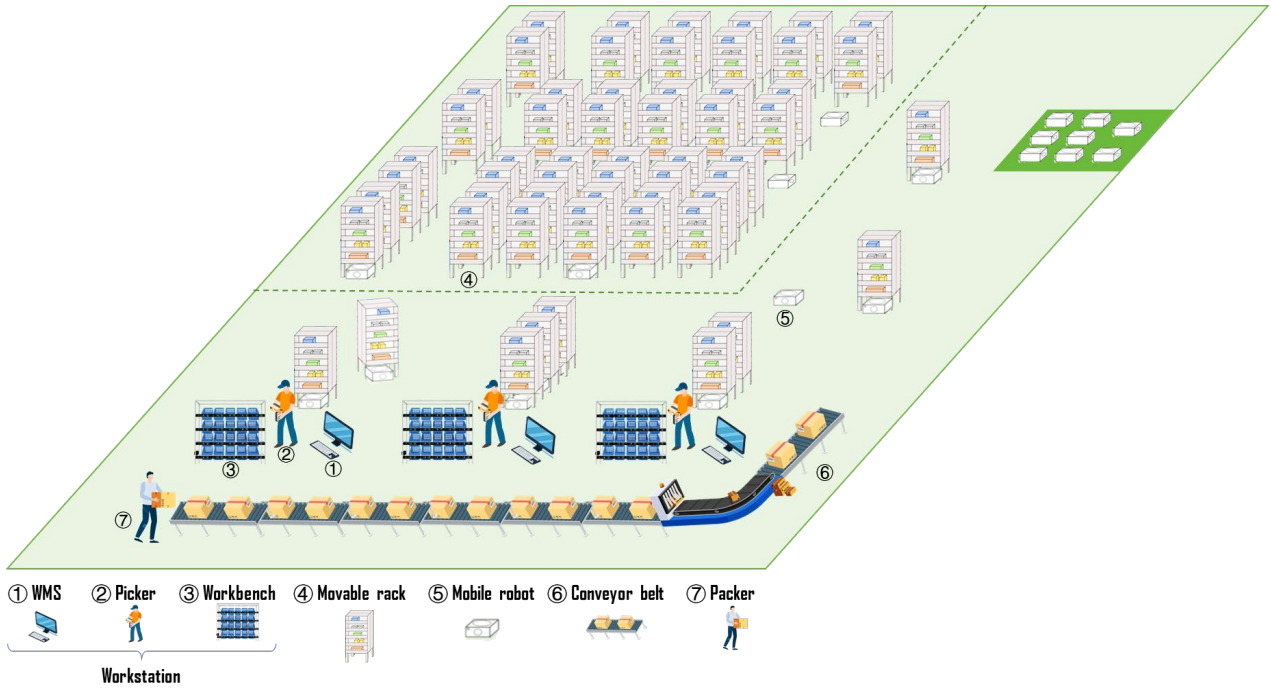

**Figure 1.** The layout of a KIVA warehouse with three workstations.

In this paper we consider the order picking process in the context of the KIVA warehouse, which is the first representative application of RMFS (Enright & Wurman, 2011). Figure 1 illustrates the layout of the KIVA warehouse, in which orders, racks, and workstations are the crucial elements that affect the picking efficiency (Weidinger et al., 2018): First, a batch of orders received by the warehousing management system (WMS) needs to be distributed to the individual workstations. Then, each workstation receives a fixed set of orders, which are processed on the workbench. In our paper each order may consist of more than one SKU, each SKU may be stored on more than one rack, and each rack contains several SKUs, i.e., an order may need to be fulfilled by multiple racks and a rack may fully or partially satisfy multiple orders. Thus, the selection of racks to fulfill the orders on the workbenches will affect the picking efficiency.

For an isolated workstation, order picking optimization is to optimize order sequencing and rack sequencing based on the determined order set and rack set, which are intertwined decisions. Weidinger et al. (2018) referred to the above problem as the *picking station scheduling* problem. Specifically, order sequencing determines the SKUs to be assembled simultaneously on the workbench, which needs to be synchronized with the racks selected to serve the station satisfying the corresponding demands, i.e., the rack disposal sequence. Optimized order sequencing and rack sequencing can improve the picking efficiency and halve the robot fleet (Boysen et al., 2017; Yang et al., 2021). However, order picking optimization in the real-world KIVA warehouse, which has multiple workstations in parallel, is more challenging than the single-workstation environment (Zhuang et al., 2021). As the scheduling problem involves both task allocation and sequencing, the problem complexity grows exponentially with the number of machines, rendering the problem intractable (Mokotoff, 2001). We can only deal with optimal order and rack sequencing in the single-workstation context, while in the multi-workstation case, it is necessary to find an optimal assignment of order-workstation along with rack-workstation.

Motivated by the above observation, we study the joint optimization of order assignment and picking station scheduling in the KIVA warehouse, seeking to address the following questions with a view to reducing the total number of rack visits: First, which orders should be assigned to the same picking station? Second, how to sort the pending orders at each station? Third, which racks and when should they be selected to satisfy the orders synchronously on every single workbench at a time, i.e., which racks should be moved to which picker, and the sequence in which the racks are presented to the picker?

**1.2  Contributions and paper organization**

Nowadays, increasing numbers of similar robotic systems, which markedly affect the manual picking decision, have been installed to support modern B2C distribution centres. Despite the widespread use of this type of system in real life, it has not received comparable research attention. In this paper we consider joint optimization in the order picking process in the KIVA warehouse, which assigns orders to multiple stations while determining the processing sequence of the orders and synchronization of the racks at the pickers. We make three contributions in this paper as follows:

First, we develop a discrete-time mathematical model for the intractable problem under study while accounting for workload balancing, which can find the optimal solutions for small-sized instances. To the best of our knowledge, this is the first study to consider order assignment in order picking optimization in the multi-workstation setting. Second, showing that the problem is *NP*-hard, we propose a hybrid search approach based on simulated annealing (SA) and beam search (BS), which we show through numerical studies provides good quality solutions for real-world-sized instances. Furthermore, we compare our approach with a rule-based greedy method and a separate optimization process without improving the order assignment policy. Our approach consistently outperforms the other methods, highlighting the necessity of joint optimization of interlinked processes. Third, we explore the potential benefits of adjusting important configuration parameters based on the warehouse layout, which provide managerial insights for improving picking efficiency to some extent.

We organize the rest of the paper as follows: In Section 2 we review the related literature to identify the research gaps and position our paper. We discuss the practical workflow of the KIVA warehouse and introduce the problem in Section 3. In Section 4 we formulate the problem as a mathematical model to treat the general problem. We present the proposed heuristic procedure in Section 5. We report the results of numerical studies conducted to assess the proposed approach and discuss the managerial insights of the research findings in Section 6. Finally, in Section 7, we conclude the paper and suggest topics for future research.

## 2   Literature Review

There is an abundance of research on optimizing order processing in picker-to-parts warehouses, detailed summaries of which can be found in de Koster et al. (2007) and Gu et al. (2010). As the scale of e-commerce shipments expands, many B2C distribution centres have applied automatic systems involving a wide range of parts-to-picker technologies to release the manual labour (see, e.g., Zaerpour et al., 2017; Kumawat & Roy, 2021). The elementary decisions concerning the daily operations of a robotic warehousing system are essentially the same as those of the traditional manual warehouse. However, the autonomous rack delivery assisted by robots in the former system requires modifications of the decision-making process (Weidinger et al., 2018).

Specifically, in the traditional warehouse, the storage area is static, from which each picker carries a fixed order batch each time, walking or driving through the area to execute the task. The source of gaining efficiency often involves optimization of the zoning, order batching, batch sequencing, and picker routing decisions (Scholz et al., 2017). However, in the KIVA warehouse, thanks to the presence of mobile robots and movable racks, the storage assignment changes dynamically, orders are processed one after the other instead of in fixed batches, and the picker can reach any rack without moving, so picking efficiency will benefit from the synergetic interaction of the orders, racks, and workstations. Accordingly, there exist significant and challenging decisions concerning parts-to-picker in the KIVA system, such as layout design, storage assignment, order picking, and robotic planning, which have not been adequately addressed (Azadeh et al., 2017; Boysen et al., 2019), yielding the following research challenges.

- Most studies on layout design can be categorized as systematic analysis, focusing on modelling techniques to estimate the performance of different system scenarios without considering optimization. Lamballais et al. (2017) used queueing theory to analytically estimate the maximum order throughput, average order cycle time, and robot utilization in a robotic mobile fulfillment system. They verified the analytic estimates by simulation. They finally suggested the optimal configuration and reasonable operating policies for the warehouse manager based on their finding that the maximum order throughput is insensitive to the dimensional parameters of the storage area.

- Most studies on storage assignment consider two elementary decisions: one is which SKUs should be stored together on the same rack and the other is where the racks should be placed in the storage area. A typical storage choice in the KIVA warehouse is the mixed-shelves policy, under which the items of the same SKU are spread all over the warehouse on multiple racks (Bartholdi & Hackman, 2014). This scattered storage contributes to a greater probability that some racks holding the requested items are close by (Weidinger & Boysen, 2018). Moreover, all the racks are identical and can be re-located dynamically to any parking position. Thus, for the latter decision, Weidinger et al. (2018) addressed the problem of where to park the racks during order processing when they are consistently moved between the picking stations and the storage area.

- Most studies on robotic planning concern the task allocation and traffic planning decisions, which coordinate the mobile robots with all their different destinations and avoid deadlocks. The coordination of multiple agents has attracted most research on rack-moving mobile robots (see, e.g., Wurman et al., 2008; D'Andrea & Wurman, 2008; Roodbergen & Vis, 2009).

In this paper we focus on the order picking process, which is at the heart of any warehouse (Azadeh et al., 2017). Van Gils et al. (2018) presented a comprehensive classification and review of picking systems. Winkelhaus et al. (2021) developed a framework for Order Picking 4.0 as a socio-technical system, considering substitutive and supportive technologies. The KIVA picking system deploys mobile robots to bring movable racks to stationary pickers so that each picker concentrates only on picking station scheduling. Note that the orders always far exceed the station capacity, which implies that a batching decision should be made to concurrently process some orders on the workbench. By synchronizing the batches of orders to be jointly handled on the bench and the racks visiting a station, fewer racks may be delivered to a station (Boysen et al., 2017). There is plentiful research on order batching in the traditional warehouse, which divides orders by certain rules to achieve specified objectives (Pan et al., 2015; Çeven & Gue, 2017; Ardjmand et al., 2018). In contrast, the batches in the KIVA setting change in real time because each order may have a different processing time, which introduces a new decision on the sorting of a given set of orders dynamically according to the actual processing of the orders concerned.

Considering that orders and racks are processed in a synchronized manner, there exists a close and mutually restrictive connection among the orders, racks, and workstations. The picking station scheduling problem is called the "mobile-robots-based order picking problem" in the literature (Boysen et al., 2017), which concerns a single picker with a given set of orders to be picked from a given set of racks. Moreover, the order sequencing decision is closely coupled with rack scheduling, necessitating the choice of the most suited racks and determining their arrival sequence (Yang et al., 2021; Zhuang et al., 2021). Most studies treat the picking stations as isolated; however, in practice, multiple stations operate in parallel, and the racks assigned to each station are not known before order allocation. Therefore, how to schedule the allocation of both orders and racks to minimize the total number of rack visits becomes another important issue in the mobile-robots-based distribution warehouse (Xie et al., 2021; Valle & Beasley, 2021). Consequently, the KIVA picking process involves several interrelated decision problems, which are often addressed separately for ease of treatment.

In conclusion, there are a few studies on the problem of joint optimization of order assignment and picking station scheduling in the KIVA warehouse, i.e., simultaneously deciding the order sets and rack sets assigned to workstations and synchronizing their processing sequence. Merschformann et al. (2017) sketched the crucial relationships between our decision problems, which facilitate the exploitation of their synergy or avoidance of sabotaging one another's success. Regarding our contribution to the literature, our work falls within the operations and control domain of the structure suggested by Azadeh et al. (2017) and addresses the first two most important decisions of the four-level hierarchy for order picking proposed by Weidinger et al. (2018).

## 3   Problem Description

We consider joint optimization in the picking process of the KIVA system, which comprises decisions on order and rack allocation to workstations and disposing sequence at each workstation. We present the problem under study after introducing the robot-assisted KIVA system.

### 3.1   The robot-based KIVA order picking system

A warehousing system is defined as a combination of hardware and processes regulating the workflow among the hardware elements applied (Boysen et al., 2017).

### 3.1.1 The basic picking elements

The KIVA system considered in this paper consists of four basic elements to enable the picking function, namely movable storage racks, picking workstations, multiple pickers, and mobile robots. The mobile robots are powered by electricity, and they perform the rotation and lifting mechanisms with flexible wheels. As for their moving directions, the warehouse floor is invisibly subdivided into grids, each of which is marked with a barcode. An integrated camera system is used to continuously read the barcodes and the robot positions. The rotation mechanism makes the robot move linearly on all sides and the lifting unit can support more than 1,000 kilograms, so that the robot can complete a robotic task of moving under the rack, lifting it, and transporting it from the storage area to a workstation (D'Andrea et al., 2008). More details of the system are given in Enright & Wurman (2011).

### 3.1.2 The elementary picking workflow

The elementary picking workflow in the KIVA system is as follows: (1) WMS receives the orders and divides them into several batches. Each workstation receives a batch of fixed orders assigned by the system to be disposed of. (2) WMS simultaneously determines the racks to be allocated to a certain workstation. The selected racks must enable the workstation to satisfy all the SKUs for the assigned orders. (3) At each workstation, a static picker identifies the order bins with barcode labels and places them on the bench in turn based on the defined sequence. Whenever an order is completed, the corresponding bin is packed and moved out of the station. The vacant position is replaced by the subsequent one. (4) According to the active orders on the workbench, the allocated racks arrive in line. The picker retrieves items from the current rack and puts them in the relevant bins to satisfy each order.

### 3.2 The order assignment and picking station scheduling problem (OAPSSP)

We formulate the above order picking process in the KIVA system as follows: There is a set of $m$ workstations $P = \{1, ..., m\}, p \in P$. We define $S$ as the set of all the SKUs and $R$ as the set of all the useable racks. First, the warehouse receives a given order set $O$ consisting of $n$ orders to be processed, i.e., the wave picking policy is used. These orders are divided into $m$ sets, and each set contains an approximate average number of orders. The set of all the orders is defined by $O = O^1 \cup O^2 ... \cup O^m$, where $O^{p'} \cap O^p = 0, 1 \leq p' < p \leq m$, i.e., each order can only be processed at one workstation. Each order $o \subseteq S$ is defined as a set of SKUs required by the customer. Then we need to determine the $m$ sets of given racks, where each $R^p \subseteq R$ is the rack set allocated to $p$ in order to enable the picking of its assigned orders. Each rack $r \subseteq S$ is defined by the set of SKUs it contains. In order to form a complete solution for OAPSSP, we further introduce two parameters $\theta^p$ and $\mu^p$ to illustrate the order sequencing policy and rack scheduling policy at workstation $p$, separately.

We assume that each rack contains many units of any SKU that can satisfy the demands of all the orders. This assumption is reasonable in the real-world e-commerce distribution centre as the order size is usually small. We also assume that a rack can be allocated to different stations and rack re-visits are allowed. We discuss this assumption further in the following (see Figure 3). Moreover, the processing capacity of each workbench is $C$ bins in parallel, i.e., the number of active orders to be tackled simultaneously per station is limited to $C$. Suppose that the number of orders contained in each order set is much larger than $C$. Then it is necessary to coordinate the displayed sequences of the bins on the workbench and of the rational racks serving them over time. Note that in our setting, after finishing an order, the succeeding order, which substitutes its predecessor in the same position of the workbench, is still able to pick items from the current rack. We first provide an overall example to illustrate the picking workflow described above (see Figure 2).

*Example:* Consider a set $S = \{A, B, C, D, E, F, G\}$ of different SKUs, which are included in $n = 25$ orders distributed to $m = 5$ workstations. The capacity of each workbench is limited to $C = 3$. Although we address the problem for all the stations simultaneously, we only consider $p = 5$ in detail as an illustration as follows: It receives order set $O^5$, in which $o_5 = \{C, E\}, o_2 = \{C\}, o_4 = \{A, G\}, o_1 = \{B, C\}$, and $o_3 = \{A, B\}$ are tackled in sequence. Then three racks $r_1 = \{A, B\}, r_2 = \{C, F\}, r_4 = \{E, G\}$ form $R^5$, which are delivered to the station in the sequence $r_4, r_2, r_1$. Figure 2 depicts a solution for $(\theta^5, \mu^5)$ where all the orders are satisfied after three rack visits.

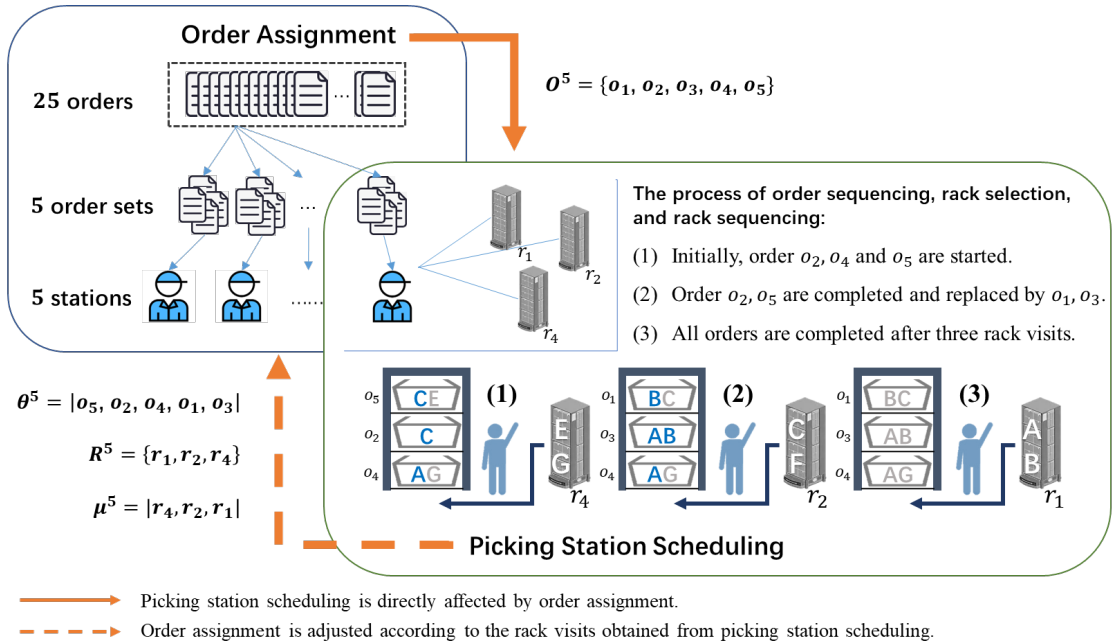

**Figure 2.** The overall example of the OAPSSP.

Furthermore, we introduce the horizon of "slot" to track the sequences of orders and racks at multiple stations, which is also used to discretize the time in our mathematical model. Each slot portrays a period in which a certain subset of the orders and a certain rack are concurrently processed at a workstation. The next slot will not emerge until one or more of the changes mentioned before occurring, i.e., two successive slots differ in at least one order being processed or in the visiting rack. For the convenience of the reader, we provide the following illustration, in which there are $m = 2$ workstations picking $n = 8$ orders with $C = 2$.

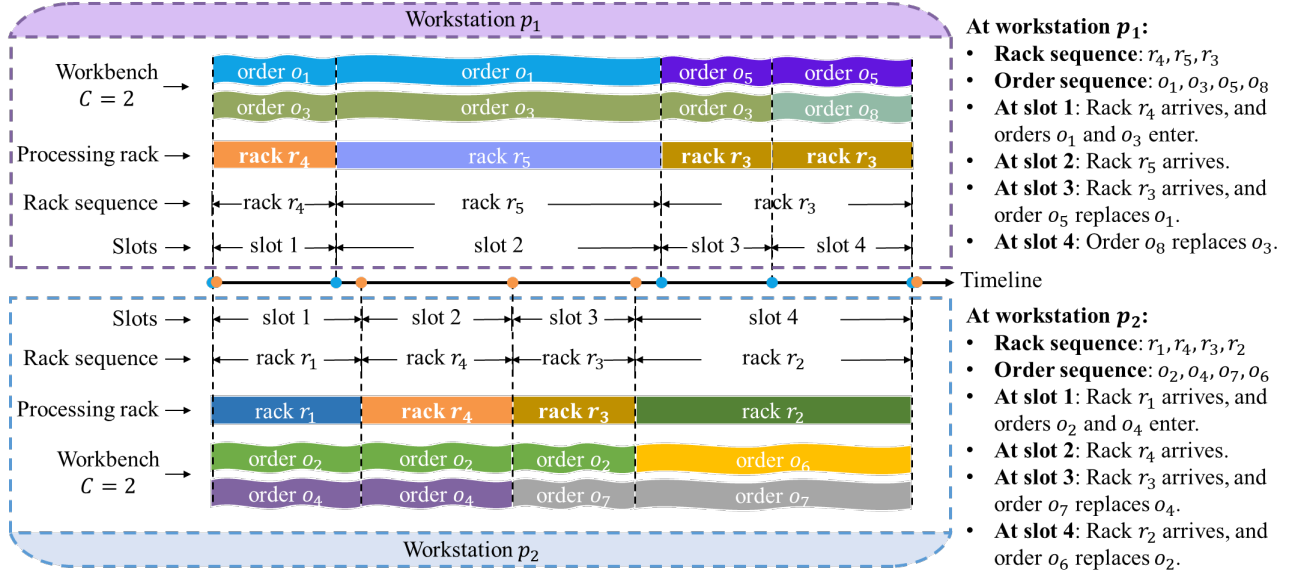

**Figure 3a.** An illustrative example of multiple workstations (without rack conflicts).

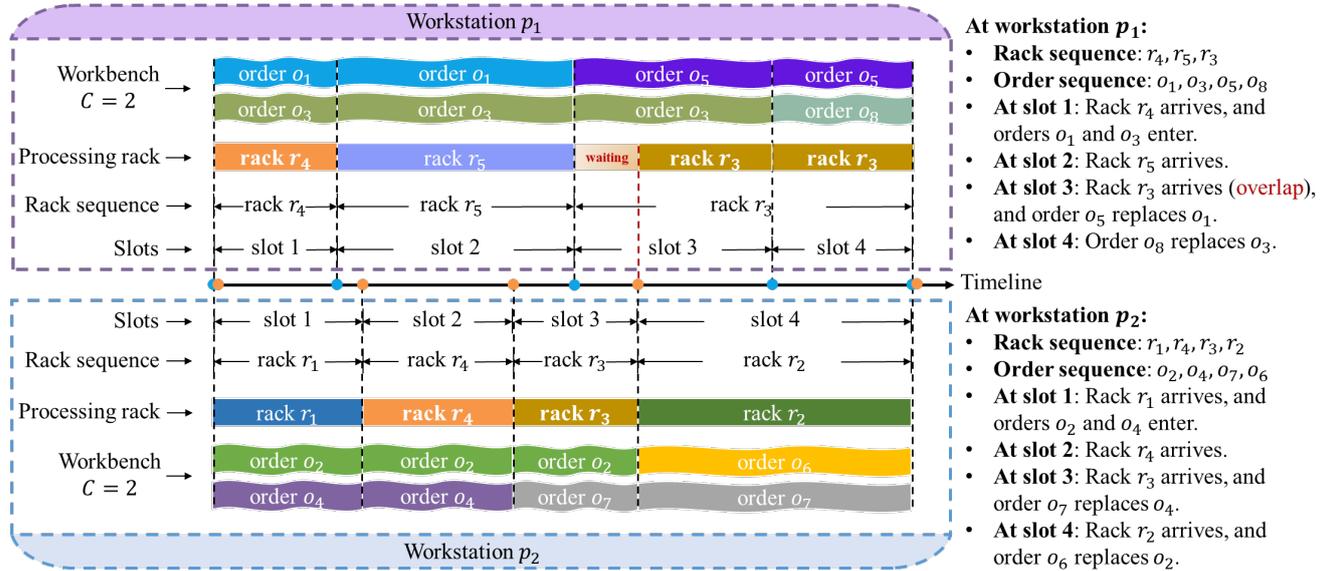

**Figure 3b.** An illustrative example of multiple workstations (with rack conflicts).

Figure 3a depicts the relationships among the slots, order and rack sequences, and time. From the perspective of workstation $p_1$, slot 1 is moved to slot 2 since the rack picked by this station is changed ($r_4 \to r_5$); then slot 2 is moved to slot 3 since the orders on the workbench and rack picked by this station are both changed ($o_5 \to o_1, r_5 \to r_3$); finally, slot 3 is moved to slot 4 since the orders on the workbench are changed ($o_3 \to o_8$). The above circle is also the case at workstation $p_2$. Recall that the same rack can be allocated to several stations over time in our setting, which may lead to time overlap of rack allocation over stations (see Figure 3b). However, the time taken by a picker to retrieve each item from a rack is negligible in practice compared with the time taken by a robot to deliver another rack from the storage area. Moreover, the picker can do alternative work during the waiting time brought by a rack change, e.g., folding a new cardboard box. We thus suppose that the waiting times due to rack conflicts are ignored in our model.

## 4 Model formulation

### 4.1 The OAPSSP model

According to the above-defined OAPSSP, we propose a comprehensive mixed-integer programming (MIP) model to minimize the total number of rack visits. We first explain why this objective is considered well-suited and fundamental for such a system as follows:

- The fewer rack visits needed to satisfy the given order set are, the smaller is the number of robots the picking process requires. For a real-world distribution centre that have many picking stations in parallel, this objective can reduce the cost of the equipment, the risk of robot collision, and even the possibility of bottleneck situations where every robot is too busy to handle a new delivering task. Boysen et al. (2017) found through simulation studies that this objective may even allow halving the fleet of robots.

- Once each order set is defined and assigned to each station, the only time component influenced by the scheduling strategy is the picker's waiting time, i.e., the time between the departure of the current rack and the arrival of the succeeding one from the storage area. In addition, the more breaks caused by rack change occurrences are, the longer is the time the picker needs to wait. Under the above circumstances, the rack visit number is the main lever for increasing the picking throughput, i.e., minimizing the visit number also minimizes the makespan of processing the given order set.

**Table 1.** Notation.

| | |
|---|---|
| ***Parameter*** | |
| $P$ | Set of all the workstations, $p \in P$ |
| $S$ | Set of all the SKUs the warehouse holds, $i \in S$ |
| $R$ | Set of all the usable racks, $r \in R$ |
| $R_i$ | Set of racks that contain SKU $i$, $R_i \subseteq R$ |
| $O$ | Set of all the orders to be processed, $o \in O$ |
| $C$ | Capacity of each workstation |
| $m$ | Number of workstations |
| $n$ | Number of orders to be processed |
| $T$ | Number of slots, $t = 1, \dots, T$ |
| $\Gamma$ | The total number of rack visits |

*Decision variable*

| | |
|---|---|
| $\alpha_t^p$ | Continuous variable: 1, if the racks visiting station $p$ in $t-1$ and $t$ differ |
| $x_o^p$ | Binary variable: 1, if order $o$ is assigned to station $p$ |
| $k_{ot}^p$ | Continuous variable: 1, if order $o$ is tackled at station $p$ in slot $t$ |
| $y_r^p$ | Binary variable: 1, if rack $r$ is assigned to station $p$ |
| $l_{rt}^p$ | Binary variable: 1, if rack $r$ arrives at station $p$ in slot $t$ |
| $\pi_{iot}^p$ | Binary variable: 1, if SKU $i$ of order $o$ tackled at station $p$ is delivered in slot $t$ |

Summarizing the notation for the problem in Table 1, we formulate our MIP model, which consists of the objective function (1) and constraints (2) to (19). For all the workstations, each one has up to $T$ slots. We can derive a trivial upper bound on the number of slots $T = \max\left\{\left\lceil\frac{n}{m}\right\rceil \cdot |o| \cdot |R|\right\}, \forall o \in O$, where $\left\lceil\frac{x}{y}\right\rceil$ means rounding the result of $\frac{x}{y}$ upwards, while $\left\lfloor\frac{x}{y}\right\rfloor$ means rounding the result of $\frac{x}{y}$ downwards.

$$(OAPSSP)\ Minimize\ \Gamma = \sum_{p \in P} \sum_{t=2}^{T} \alpha_t^p \tag{1}$$

subject to

$$\sum_{o \in O} x_o^p = \begin{cases} \left\lceil\frac{n}{m}\right\rceil, & 1 \leq p \leq n\%m \\ \left\lfloor\frac{n}{m}\right\rfloor, & n\%m < p \leq m \end{cases}, \quad \forall p \in P \tag{2}$$

$$\sum_{p \in P} x_o^p \leq 1, \quad \forall o \in O \tag{3}$$

$$\sum_{t=1}^{T} k_{ot}^p \leq T \cdot x_o^p, \quad \forall o \in O, \forall p \in P \tag{4}$$

$$\sum_{t=1}^{T} k_{ot}^p \geq x_o^p, \quad \forall o \in O, \forall p \in P \tag{5}$$

$$\sum_{o \in O} k_{ot}^p \leq C, \quad \forall t = 1, \dots, T, \forall p \in P \tag{6}$$

$$\sum_{r \in R} l_{rt}^p \leq 1, \quad \forall t = 1, \dots T, \forall p \in P \tag{7}$$

$$\sum_{t=1}^{T} l_{rt}^p \leq T \cdot y_r^p, \quad \forall r \in R, \forall p \in P \tag{8}$$

$$\sum_{t=1}^{T} l_{rt}^p \geq y_r^p, \quad \forall r \in R, \forall p \in P \tag{9}$$

$$\sum_{t=1}^{T} \pi_{iot}^{p} \geq x_{o}^{p}, \qquad \forall i \in o, \forall o \in O, \forall p \in P \qquad (10)$$

$$\sum_{r \in R_i} l_{rt}^{p} + k_{ot}^{p} \geq 2\pi_{iot}^{p}, \qquad \forall i \in o, \forall o \in O, \forall p \in P, \forall t = 1, \ldots, T \qquad (11)$$

$$k_{ot}^{p} + k_{ot'}^{p} \leq k_{ot''}^{p} + 1, \qquad \forall o \in O, \forall p \in P, 1 \leq t < t'' < t' \leq T \qquad (12)$$

$$l_{rt}^{p} - l_{r(t-1)}^{p} \leq \alpha_{t}^{p}, \qquad \forall r \in R, \forall p \in P, \forall t = 2, \ldots, T \qquad (13)$$

$$0 \leq \alpha_{t}^{p} \leq 1, \qquad \forall p \in P, \forall t = 1, \ldots, T \qquad (14)$$

$$0 \leq k_{ot}^{p} \leq 1, \qquad \forall o \in O, \forall p \in P, \forall t = 1, \ldots, T \qquad (15)$$

$$x_{o}^{p} \in \{0,1\}, \qquad \forall o \in O, \forall p \in P \qquad (16)$$

$$y_{r}^{p} \in \{0,1\}, \qquad \forall r \in R, \forall p \in P \qquad (17)$$

$$l_{rt}^{p} \in \{0,1\}, \qquad \forall r \in R, \forall t = 1, \ldots, T, \forall p \in P \qquad (18)$$

$$\pi_{iot}^{p} \in \{0,1\}, \qquad \forall i \in o, \forall o \in O, \forall t = 1, \ldots, T, \forall p \in P \qquad (19)$$

Objective (1) minimizes the total number of rack visits at all the workstations, which is modelled as the number of times rack switching takes place at workstations. Eq. (2) ensures that the workload assigned to each workstation is approximately balanced, where $n\%m$ means the modulus operation. Note that any unbalanced workload assignment may lead to lower order picking efficiency due to the idle time of pickers. In this paper the number of orders is used to measure the workload, which is easily implemented (Valle & Beasley, 2021) and has been shown to provide yield good performance for workload balancing (Zhuang et al., 2021). Eq. (3) guarantees that each order is allocated to only one workstation. Eqs (4) and (5) ensure that only if an order is allocated to a certain workstation will it be processed at the station in some slot(s) or it will not be processed at the station at all. Eqs (6) and (7) guarantee that in each slot, at most $C$ orders are processed or at most one rack is visiting, which reflects the capacity limit of each workbench. Eqs (8) and (9) ensure that only if a rack is assigned to a certain workstation, will it be processed at the station in some slot(s) or it is not processed at the station at all. Eq. (10) states that all the SKUs required by order $o_i$ allocated to workstation $p$ should be delivered at the station, which can happen only in a slot where both $o_i$ and a suitable rack are concurrently processed due to Eq. (11). Eq. (12) guarantees that an order must be processed in succession. Finally, Eq. (13) records the rack visiting changes. Eqs (14) to (19) are the integrality constraints, i.e., the domain restrictions of the decision variables. Note that $\alpha_t^p$ and $k_{ot}^p$ are forced to take either the value 1 or 0 due to Eqs (12) and (13), and the binary nature of the other variables.

The complexity of the proposed MIP model depends on the numbers of customer orders, required SKUs, feasible racks, and workstations. The larger the instance size becomes, the larger is the solution space, as well as the number of constraints. Therefore, solving the MIP model by a commercial solver such as Gurobi is very difficult and time consuming. In addition,

OAPSSP is not only a complicated MIP model but also an *NP*-hard problem because if the orders and racks allocated to each workstation are fixed, it reduces to the mobile robot-based order picking problem (MROP). Boysen et al. (2017) showed that MROP is *NP*-hard by reducing it to the set covering problem (Garey & Johnson, 1979). Consequently, efficient heuristic algorithms are necessary for tackling the problem, the development and implementation of which we discuss in the next section.

**4.2 A lower bound derived from a relaxation model**

Consider the special case that does not account for the temporal dynamics of the OAPSSP problem, i.e., the orders and racks are assigned to workstations just in a way that the racks are sufficient to serve all the assigned orders. In other words, this case reflects the situation where the capacity of each workstation is unlimited so that it can process all the assigned orders simultaneously. Note that the objective in this case is to minimize the total number of racks allocated to workstations $\Lambda$.

In the following we propose a MIP formulation to reflect the above order and rack allocation problem, denoted as ORAP. We add the binary variable $\delta_{io}^p$ to denote if SKU $i$ of order $o$ is processed at workstation $p$ or not. There are two sets of decision variables $y_r^p$ and $x_o^p$ to indicate if rack $r$ or order $o$ is needed by workstation $p$ or not, respectively. We formulate the resulting problem as follows:

$$(ORAP)\ Minimize\ \Lambda = \sum_{p \in P} \sum_{r \in R} y_r^p \qquad (20)$$

subject to

$$\sum_{o \in O} x_o^p = \begin{cases} \left\lceil \frac{n}{m} \right\rceil, & 1 \leq p \leq n\%m \\ \left\lfloor \frac{n}{m} \right\rfloor, & n\%m < p \leq m \end{cases}, \quad \forall p \in P \qquad (21)$$

$$\sum_{p \in P} x_o^p \leq 1, \qquad \forall o \in O \qquad (22)$$

$$\delta_{io}^p \geq x_o^p, \qquad \forall i \in o, \forall o \in O, \forall p \in P \qquad (23)$$

$$\sum_{r \in R_i} y_r^p + x_o^p \geq 2\delta_{io}^p, \qquad \forall i \in o, \forall o \in O, \forall p \in P \qquad (24)$$

$$x_o^p \in \{0,1\}, \qquad \forall o \in O, \forall p \in P \qquad (25)$$

$$y_r^p \in \{0,1\}, \qquad \forall r \in R, \forall p \in P \qquad (26)$$

$$\delta_{io}^p \in \{0,1\}, \qquad \forall i \in o, \forall o \in O, \forall p \in P \qquad (27)$$

Objective (20) minimizes the total number of racks allocated to process all the pending orders. Eq. (21) is the workload balance constraint. Eq. (22) guarantees that each order is allocated to only one workstation. Eq. (23) states that all the SKUs required by order $o_i$ allocated to workstation $p$ should be delivered to the station, which only happens when both conditions hold, i.e., order $o$ and some rack containing its required SKUs are assigned to the special workstation due to Eq. (24).

## 5 The Proposed Approach

As discussed above, applying standard solvers, e.g., Gurobi, can only solve OAPSSP for small instances, which becomes computationally inefficient to deal with real-life-scale problems. Thus, in this section, we provide a heuristic algorithm to search for approximate solutions within practically suitable running times.

### 5.1 The solution search framework

Given a set of multi-SKU orders, a set of racks each containing a subset of SKUs, and several workstations, OAPSSP is concerned with deciding: (a) which orders to assign to each workstation; (b) in what sequence each workstation should process the orders, where up to a certain number of orders can be processed simultaneously; and (c) which rack should be sent to which workstation in each time slot. Note that order sequencing and rack scheduling are interlinked decisions, where no matter which one is given, the other one remains *NP*-hard (Boysen et al., 2017). We apply a search framework that constructs solutions and improves them in steps as follows:

- A good initial solution is obtained in two steps. We first solve the relaxation model via a standard solver, according to which the order set and rack set assigned to each workstation are given. Then the problem reduces to the order sequencing and rack sequencing problem with $m$ independent workstations. The initial solution can thus be determined in a random manner, i.e., given a random permutation of the orders assigned to each workstation, the synchronously suitable racks are selected in succession.

- The initial solution is then iteratively improved based on SA, a randomized search scheme (see Section 5.3). Since a complete solution contains two elements related to order and rack, each neighbourhood solution is also generated in two steps. First, we propose several neighbourhood operators, one of which is applied to simultaneously generate $m$ order sequences for all the workstations as a half part of the neighbourhood solution. Second, for given decisions of orders, each rack schedule is decided in a heuristic fashion, employing a version of best first search where only a subset of the alternatives is explored in each branch (Beam Search, see Section 5.2).

### 5.2 Rack scheduling policy for isolated workstations

For each isolated workstation, one solution $x^p = (\theta^p, \mu^p)$ consists of the order processing policy $\theta^p = |o_1^p, o_2^p, ...|$ and the rack scheduling policy $\mu^p = |r_1^p, r_2^p, ...|$. The decisions of racks $\mu^p$ is completed with a fixed $\theta^p$ that heuristically minimizes the number of rack visits. We derive $\mu^p$ from a designed BS procedure, which is based on dynamic programming scheme.

We briefly introduce the general BS mechanism. A graph search heuristic initially applied in the field of speech recognition, BS was first introduced to solve scheduling problems and compared with other well-known heuristics by Ow & Morton (1988). Since then, BS has been extended to a powerful metaheuristic applicable to many real-world optimization problems (Blum, 2005; Boysen & Zenker, 2013). More details of the BS heuristic and its extensions can be found in Sabuncuoglu et al. (2008). BS executes the searching procedure based on a tree representation of the solution. However, it does not apply the breadth-first approach (e.g., the branch-and-bound technique) but restricts the number of nodes per stage to be further branched to a promising subset. The size of the subset is determined by a given parameter, i.e., the beam width $BW$, and the nodes to be selected in the subset are evaluated by a filtering process. Thus, the search process can be illustrated as follows: Starting with the root node, all the nodes of stage 1 are built, among which the promising nodes are identified by filtering. Note that filtering can be obtained by a priority value based on a specific issue. Thus, the promising subset of stage 1 consists of the $BW$ best nodes found by filtering, which are further branched to construct the set of nodes in stage 2. Then again, filtering is applied to delete some poor nodes of stage 2. The above steps continue until the final stage is reached and the result of BS is returned.

Three components need to be pre-defined when applying BS for a specific problem, namely the DP structure, beam width, and filtering rule. In the following we provide their specifications for our problem.

- DP structure

We introduce the DP procedure that can be directly used for BS. The procedure can be subdivided into no more than $T + 1$ stages $s^p = 0, \dots, T$, where stage $s^p = 0$ is defined as the initial stage and $s^p = 1, \dots, T$ determines the allocation of a specific rack to the $s^p$th position of $\mu^p$. Note that $T$ corresponds to the upper bound on the number of rack visits analyzed in Section 4. Each stage $s^p$ contains a joint state $(\tilde{o}_1^p, \dots, \tilde{o}_C^p, \psi^p, s^p)$ including three elements:

- the initial stage holds $(o_1^p, \dots, o_C^p, C + 1, 0)$;
- $\tilde{o}_c^p$ represents the set of unsatisfied SKUs required by the order currently processed in the $c$th space position on workbench $c = 1, \dots, C$;
- pointer $\psi^p$ refers to the next serial number of the order to be processed in the given order sequence $\theta^p$.

As mentioned before, once an order is processed, it need not wait for the arrival of the succeeding rack. Moreover, the SKUs in the order should be provided from the first available rack containing it when the order is active on the workbench. Then we specify the transitions of the state elements, which exist only between two coterminous stages, as follows: There are two possible transitions when assigning rack $j$ to the $(s^p + 1)$th position of $\mu^p$ based on the current state $(\tilde{o}_1^p, \dots, \tilde{o}_C^p, \psi^p, s^p)$:

- if $\tilde{o}_c^p \backslash r_j \neq \emptyset$ for each $c$, then the state changes to $(\tilde{o}_1^p \backslash r_j, \ldots, \tilde{o}_C^p \backslash r_j, \psi^p, s^p + 1)$;
- if $\tilde{o}_c^p \backslash r_j = \emptyset$ for any $c$ (may be more than one, supposed as $\bar{C}$), i.e., there exist $\bar{C}$ ($\bar{C} \leq C$) positions on the workbench in which the orders can be fully satisfied by rack $j$, then the pending orders in $[\psi^p, \ldots, \psi^p + \bar{C} - 1]$ positions of $\theta^p$ substitute these orders. Consequently, the corresponding $\tilde{o}_c^p$ change to $o_{\psi^p}^p \backslash r_j, \ldots, o_{\psi^p + \bar{C} - 1}^p \backslash r_j$, respectively, while $\psi^p$ changes to $\psi^p + \bar{C}$ and $s^p$ changes to $s^p + 1$.

Finally, the succeeding stages will not terminate until the state $(\tilde{o}_1^p = \emptyset, \ldots, \tilde{o}_C^p = \emptyset, |O^p| + 1, s^{p*})$ is reached, which represents that the order picking process at workstation $p$ is finished. Furthermore, for each workstation, the optimal objective value is equal to $s^{p*}$ and the optimal rack sequence $\mu^{p*}$ can be simply obtained based on backward recursion.

- Beam width

There may exist a poor upper bound $UB$ when applying a comparatively large $BW$, which hurts the performance of BS. Thus, an iterated beam search (*IBS*) can be applied to make BS benefit from a tight $UB$. Specifically, we initialize an ordered increasing number of beam widths $BW$s. First, BS is executed with a small $BW$ for quickly generating an initial $UB$, which is passed to the next iteration of BS executed with a larger $BW$ and so on. We set $\gamma = 10$ in the numerical studies.

---

**Algorithm** *IBS*

1: Input: the maximum number $\gamma = 10$ of $BW$s
2: Initialize $UB = \infty$ and $i = 1$
3: **while** $i$ does not reach $\gamma$ **do**
4:     Solve BS with $UB$ and $BW = i$
5:     Update $UB$ = result of BS
6:     Update $i = i + 1$
7: **end while**
8: Calculate $s^{p*}$ based on the optimal sequence
9: **return** $s^{p*}$

---

- Filtering rule

Each alternative branch from one node corresponds to a different rack selection, and these are ranked and filtered via the numbers of orders they complete and as a tiebreaker the reduction in the number of SKU's pending. BS restricts the number of branches that are further explored to the $BW$ most promising ones per node. To select the $BW$ branches, we rank them according to the increasing number of orders that have not been processed, i.e., $|O^p| - \psi^p + 1$, and apply the minimum number of the currently remaining SKUs on the workbench as the tiebreaker, i.e., $|\cup_{c=1}^C \tilde{o}_c^p|$.

## 5.3 Order assignment and sequencing policy

In this section we first propose neighbourhood operators based on a designed presentation, which can be used to obtain a solution of order assignment and sequencing. Then, the solution is improved according to the SA scheme.

5.3.1 Neighbourhood construction

We denote a feasible solution as $X = (\Theta, M)$, where $\Theta = |\theta^1, \ldots, \theta^m|$ and $M = |\mu^1, \ldots, \mu^m|$. The whole order scheduling policy $\Theta$, which combines decisions of order assignment among multiple workstations and order sequencing at each one, is represented by a tuple, where the assigned sets of orders are separated by a symbol, e.g., the symbol *zero* (see Figure 4).

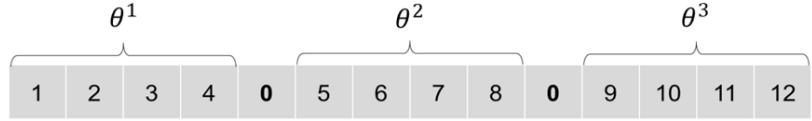

**Figure 4.** Order schedule $\Theta$ representation.

We then employ three types of neighbourhood structures $n() \in N$ based on the above representation. As mentioned before, when SA attempts to move to a new solution, one of these types is selected randomly with equal probability. Position-based neighbourhoods are commonly used for permutations that represent scheduling decisions. Therefore, we construct three position-based neighbourhood operators for the problem.

・ *Swap operator*: Select two points at random and swap their positions (see Figure 5a).

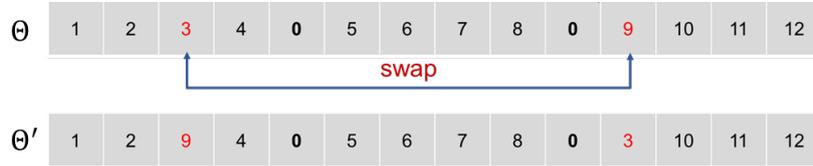

**Figure 5a.** Neighbourhood operator $n(1)$.

・ *Shift operator*: Randomly select three points and shift the points between the first two points to after the third point (see Figure 5b).

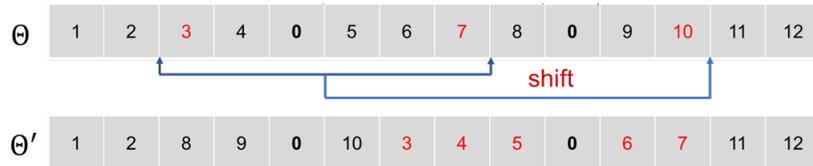

**Figure 5b.** Neighbourhood operator $n(2)$.

・ *Inversion operator*: Randomly select two points and reverse the order between them completely (see Figure 5c).

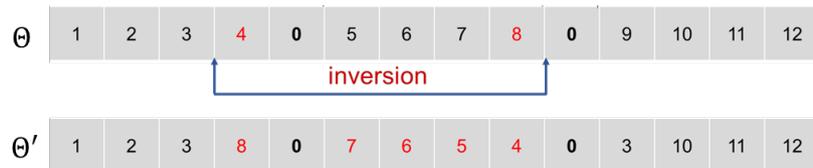

**Figure 5c.** Neighbourhood operator $n(3)$.

5.3.2 Simulated annealing scheme

SA, introduced by Kirkpatrick et al. (1983), is an algorithmic approach to solve combinatorial optimization problems (Cerný, 1985; Aarts et al., 1997). It randomizes the local search procedure and accepts changes that worsen the solution with some probability. Thus, SA constitutes an attempt to reduce the probability of getting trapped in a suboptimal solution.

Given a feasible solution $X$ of OAPSSP, $X$ is evaluated with the equation $f(X) = \sum_{p=1}^{m} s^p$, where $s^p$ is the number of rack visits at workstation $p$. During the search process, SA attempts to transform the current solution into a new one $X'$ by changing $\Theta$. In particular, the new $\Theta'$ is obtained from the neighbourhood. Once $\Theta'$ is given, the problem to obtain $X'$ reduces to the decisions of racks, which can be solved by the method proposed in Section 5.2. The new feasible solution $X'$ is always accepted if its objective value $f(X')$ is better than that of $X$, i.e., $\Delta f = f(X') - f(X) < 0$. Otherwise, $X'$ substitutes $X$ with probability $p = \exp(-\Delta f/\tau)$, where $\tau$ is a control parameter called temperature.

At the beginning of execution, $\tau$ is assigned an initial value $\tau_0$. As suggested by Ropke & Pisinger (2006) and Masson et al. (2013), a well-chosen $\tau_0$ depends on the problem instance at hand. Following them, we calculate $\tau_0$ according to our initial solution instead of specifying it as a parameter. First, we denote $\bar{f}(X) = f(X)/m$ as a modified objective function. The initial temperature is then set in such a way that $\tau \coloneqq -w \cdot \bar{f}(X_0)/ln0.5$, i.e., a solution that is $w\%$ worse than $\bar{f}(X_0)$ has a 50% probability to be accepted. Note that the modification aims to avoid a too-large value of the initial temperature. Through preliminary tests, we set $w = 0.05$, which leads to outstanding results.

At each temperature value $\tau$, the current solution iterates $K$ times, which are seen as "one epoch". According to Cho et al. (2005), the length of the initial epoch is set as 10 and modified as $K \coloneqq K + \left\lfloor K \cdot \left(1 - \exp\left(\frac{f^{min} - f^{max}}{f^{max}}\right)\right)\right\rfloor$ after each epoch. Note that $f^{min}$ is the smallest objective value recorded in the past epoch and $f^{max}$ is the largest. The temperature is lowered to $\tau \coloneqq \alpha\tau$ after each epoch, where $\alpha$ called the cooling rate is set as 0.9. Finally, we set the terminating criterion: $\tau$ falls below 0.01 or a given time frame is exhausted.

---

**Algorithm** *SA*

1: Construct an initial solution $X = (\Theta, M)$

2: Set $T = -0.05 \cdot \bar{f}(X)/ln0.5$, $K = 10$

3: Set $X^* = X$

4: **while** the stop criterion is not met **do**

5:    Initialize $X^{min}$, $X^{max}$, and $k = 1$

6:    **while** $k < K$ **do**

---

7:      Randomly choose a neighbourhood operator $n() \in N$
8:      Apply $n()$ to $\Theta$, resulting in $\Theta'$
9:         Calculate $X', f(X')$ based on $\Theta'$
10:     **if** $f(X') < f(X)$ **then**
11:        Set $X = X', X^{min} = X'$
12:     **else**
13:        Set $X = X'$ with probability $p = exp(\frac{f(X')-f(X)}{T})$
14:        Set $X^{max} = X'$
        end if
15:     **if** $X'$ is better than $X$
           Set $X^* = X'$
        end if
16:    $k = k + 1$
17:    **end while**
18:    $K = K + \left\lfloor K \cdot \left(1 - exp\left(\frac{f(x^{min})-f(x^{max})}{f(x^{max})}\right)\right)\right\rfloor$
19:    $T = 0.9T$
20: **end while**
21: **return** $X^*$

## 6 Numerical Studies

In this section we present the results of numerical studies conducted to assess the performance of our proposed heuristic approach. First, we discuss in Section 6.1 how we generate the test instances, as there exists no established testbed for OAPSSP. Then, we present in Section 6.2 the performance of our heuristic solution approaches in comparison with that of Gurobi and other designed benchmarks. Specifically, we introduce a traditional policy that is widely used for the type of warehouses under study, and another policy that executes picking station scheduling with random order assignment. Finally, we provide in Section 6.3 managerial insights and suggestions concerning the real-life robotic mobile fulfillment system according to a series of sensitivity analyses.

We conduct all the numerical studies on a 64-bit PC with an Intel Core i7-10510U (1.80GHz and 2.30GHz) and 16.0 GB main memory operating under Windows 10. We code the procedures in C++ (Visual Studio 2019) and use the off-the-shelf solver Gurobi (version 9.1.0) to solve the MIP models.

### 6.1 Instance generation

The KIVA system is typically applied in intelligent distribution centres where many small SKUs are stored in a scattered manner on racks. Our instance generation follows the real-world operating rules adopted in Jing Dong Asia No. 1 Warehouse that uses a similar system (Neuhub, 2019).

Over the planning horizon, a total of 20, 50, 500, and 1500 orders requiring to be fulfilled, where $n = 20, 50$ processed by $m = 2$ workstations represent the small-sized instances, and $n = 500, 1500$ processed by $m = 5$ workstations represent the large-sized instances. Since the number of racks significantly influences the scale of the OAPSS problem, we set the number of racks $|R|$ used to fulfill the orders to either $|R| = 10, 20, 50$ for small-sized instances and $|R| = 1,500$ for large-sized instances. Note that the warehouse must contain more racks; however, only a subset of these will be required during a given picking period. Each rack may have up to 50 storage locations (see CNN Business, 2018). Valle & Beasley (2021) assumed that each rack can store 25 different products per rack. Accordingly, we vary the number of SKUs contained in each rack $\beta = 6, 8, 10$ for small-sized instances, and $\beta = 20$ for large-sized instances, which indicates the diversity of SKUs per rack. The number of spaces on each workbench is given as $C = 3, 5$ for small-sized instances and $C = 15$ for large-sized instances.

**Table 2.** Parameters for instance generation.

| Parameter | Description | Small-sized | Large-sized |
| --- | --- | --- | --- |
| $n$ | Number of orders | 20, 50 | 500, 1000, 1500 |
| $m$ | Number of picking stations | 2 | 5 |
| $|R|$ | Number of useful racks | 10, 20, 50 | 1000 |
| $C$ | Capacity per workbench | 3, 5 | 15 |
| $\beta$ | Storage per rack | 6, 8, 10 | 20 |

Furthermore, the average order comprises 1.6 items (Boysen et al., 2019) and the vast number of orders contain only one or two items (Weidinger, 2018). Accordingly, we set the number of SKUs in each order to a random integer uniformly distributed from [1, 3]. Then, SKUs in real-world distribution centres have varying picking frequencies, i.e., some belong to the best-selling products, which take a much larger share of the total flow than the others. To consider this, we make the following rule to generate the SKUs required by each order, which are randomly selected via an exponential distribution with an exponent of 0.5 (referred to Boysen et al., 2017; Yuan et al., 2018). Each SKU contained on a rack is also selected according to the same exponential distribution with an exponent of 0.5, so the better-selling items are more likely to appear on several racks concurrently. Note that it is possible that some required SKUs may not be present in any of the racks due to the above generating rule. To prevent such infeasible instances, any remaining SKUs are added to a randomly selected rack. For reference, we summarize the parameters used for instance generation in Table 2.

### 6.2 Algorithmic performance

6.2.1 Comparison with a standard solver for small instances

In this section we test the small-sized instances and provide the results when they are solved by the standard solver Gurobi and our proposed heuristic algorithm. We propose 36 small-sized parameter settings according to different combinations of number of stations $m$, number of orders $n$, number of racks $|R|$, capacity of each workbench $C$, and storage diversity of each rack $\beta$. For each setting, we generate ten random instances based on the designed distribution described above and each instance is tested five times. We set a time limit equivalent to 1,800 CPU seconds for Gurobi.

**Table 3.** Numerical results with $m = 2$ picking stations.

| $n$ | | | 20 | | | | | | | | 50 | | | | | |
|---|---|---|---|---|---|---|---|---|---|---|---|---|---|---|---|---|
| $C$ | | 3 | | | | 5 | | | | 3 | | | | 5 | | |
| | Gurobi | | SA | | Gurobi | | SA | | Gurobi | | SA | | Gurobi | | SA | |
| $\beta$ | Π | $T$ | ΔΠ | $T$ | Π | $T$ | ΔΠ | $T$ | Π | $T$ | ΔΠ | $T$ | Π | $T$ | ΔΠ | $T$ |
| | | | | | | | $\|R\|=10$ | | | | | | | | | |
| 6 | 9.88 | 62.12 | 7.09% | 4.80 | 7.29 | 395.02 | 12.57% | 7.79 | 32.20 | 1800 | **-9.92%** | 86.09 | 27.77 | 1800 | **-6.03%** | 58.55 |
| 8 | 9.87 | 60.95 | 2.59% | 2.76 | 7.10 | 245.87 | **0.00%** | 5.34 | 31.01 | 1800 | **-2.88%** | 88.98 | 24.30 | 1800 | **-9.11%** | 39.37 |
| 10 | 8.62 | 57.26 | **0.00%** | 2.09 | 7.10 | 195.55 | **0.00%** | 4.44 | 27.88 | 1800 | **-5.09%** | 83.20 | 23.66 | 1800 | **-14.00%** | 31.33 |
| | | | | | | | $\|R\|=20$ | | | | | | | | | |
| 6 | 9.12 | 601.10 | 8.77% | 17.24 | 12.00 | 900.53 | 7.11% | 40.34 | - | 1800 | (30.19) | 104.22 | - | 1800 | (25.19) | 121.11 |
| 8 | 8.90 | 616.35 | 2.90% | 16.33 | 10.88 | 792.80 | 2.20% | 48.87 | - | 1800 | (29.10) | 92.38 | - | 1800 | (25.64) | 116.46 |
| 10 | 8.67 | 609.33 | 1.51% | 12.81 | 10.43 | 718.86 | **0.00%** | 40.33 | - | 1800 | (26.83) | 94.09 | - | 1800 | (23.12) | 107.47 |
| | | | | | | | $\|R\|=50$ | | | | | | | | | |
| 6 | 12.87 | 1800 | **-23.30%** | 58.65 | - | 1800 | (12.39) | 148.33 | - | 1800 | (25.34) | 183.04 | - | 1800 | (22.14) | 209.50 |
| 8 | 12.22 | 1800 | **-32.00%** | 59.21 | - | 1800 | (10.88) | 153.29 | - | 1800 | (22.89) | 183.78 | - | 1800 | (21.08) | 202.12 |
| 10 | 11.02 | 1800 | **-27.08%** | 54.55 | - | 1800 | (9.76) | 129.39 | - | 1800 | (22.08) | 136.34 | - | 1800 | (19.09) | 236.90 |

Table 3 presents the comparison results for small-sized cases with two picking stations. Column Π represents the optimal objective value returned from Gurobi, averaged over the instances it can solve. Column ΔΠ represents the percentages of relative optimality gaps between the best solution found by SA and the solution found by Gurobi. Column $T$ represents the average runtime, which is in seconds. We draw the following conclusions from the test results.

- Among these cases, there are five twelfth ones where Gurobi cannot find a solution within the given time frame, the objective values of which are shown as "-". The corresponding values of ΔΠ are thus incalculable, the spaces of which are filled by the objective values solved by SA (in parentheses). For the rest of the cases, we can find that the average optimality gap of SA is small: it can reach the optimal solutions in some ones. When the scale of the problem becomes large with more orders to be processed and more racks to be selected, the proposed heuristic can find a better solution than the standard solver within an acceptable runtime, and the solver is eventually not able to yield a feasible solution within the set time frame.

- Providing more selectable racks and expanding the storage density within each rack can improve the quality of the solution, i.e., enhancing picking efficiency in real-life operations, as it better utilizes the synergy between orders and racks. However, this will lead to an increase in solution time, so in practical operations, how to design shelf space and allocate shelf positions requires further trade-offs.

We also conclude that Gurobi is unable to solve real-world-sized instances of OAPSSP. Moreover, while the proposed solution approach performs well for small-sized instances, further assessment of its performance in tackling real-world-sized instances is needed.

6.2.2 Comparison with other heuristics in large instances

In this section we explore how SA performs for real-life-sized instances, for which applying Gurobi is out of the question for its lack of computing power. We introduce two comparative heuristics, namely a rule-based greedy method (denoted as RBG) widely used in the real world (Yang et al., 2021; Qin et al., 2021), and a decomposition-based interactive method (denoted as DBI) proposed by Boysen et al. (2017). Moreover, we compare the optimality gaps between our designed method and other two heuristics according to the lower bound (denoted as LB) derived from the relaxed model proposed in Section 4.2 solved by Gurobi. We also set a time limit equivalent to 1,800 CPU seconds for all the methods.

The most widely applied RBG in practice is greedy rack scheduling combined with first-come-first-served (FCFS) order processing. Specifically, according to their arriving times, the orders in the pool have a natural permutation. Then the first order is assigned to station one, the second order assigned to station two, and so on until all the orders have been allocated. For each station, the orders are released to the workbench sequentially according to the listing sequence. Then the selected racks are delivered successively, i.e., one by one, each of which contains mostly the items required by the currently active orders. Once an order is processed, it is substituted by the next order stipulated on the list, and additional racks will arrive until the whole picking process is completed.

The DBI proposed by Boysen et al. (2017) shows its efficiency when solving the special case of our problem, which focuses on each single workstation. We extend their method to the multi-station environment by fixing the orders assigned to each workstation according to the FCFS rule and compare ours to it. As the added order allocation mechanism is too straightforward to contribute to the optimization process, the improved DBI can be seen as a comparative approach that optimizes picking station scheduling separately without considering order allocation.

**Table 4.** Comparison results under different methods for the large-sized cases.

| $n$ | $m$ | $|R|$ | $C$ | $\beta$ | LB | | RBG | | | DBI | | | Our method | | |
|---|---|---|---|---|---|---|---|---|---|---|---|---|---|---|---|
| | | | | | $\Pi$ | $T$ | $\Pi$ | $\Delta\Pi$ | $T$ | $\Pi$ | $\Delta\Pi$ | $T$ | $\Pi$ | $\Delta\Pi$ | $T$ |
| 500 | 5 | 500 | 15 | 20 | 169.88 | 13.65 | 303.95 | 78.92% | 1.87 | 255.21 | 50.23% | 249.32 | 206.52 | 21.57% | 376.01 |
| | | 1000 | | | 153.29 | 13.77 | 302.81 | 97.54% | 2.03 | 244.76 | 59.67% | 310.23 | 197.84 | 29.06% | 460.00 |
| 1000 | 5 | 500 | 15 | 20 | 332.10 | 24.50 | 677.78 | 104.09% | 2.98 | 561.18 | 68.98% | 359.03 | 432.46 | 30.22% | 498.70 |
| | | 1000 | | | 298.37 | 27.09 | 647.08 | 116.87% | 3.15 | 546.82 | 83.27% | 422.72 | 417.33 | 39.87% | 543.02 |
| 1500 | 5 | 500 | 15 | 20 | 502.39 | 47.92 | 1108.57 | 120.66% | 3.22 | 992.52 | 97.56% | 590.44 | 699.43 | 39.22% | 622.32 |
| | | 1000 | | | 419.23 | 48.11 | 1033.19 | 146.45% | 5.23 | 851.45 | 103.10% | 633.24 | 621.68 | 48.29% | 698.08 |

Tables 4 presents the algorithmic performance of the three methods with five picking stations, where some findings according to the comparison results are concluded as follows:

- We first compare the results of our method with LB. The average gap between LB and ours is about 34.71%, which is moderate and implies that the real gap between our heuristic method and the optimal solution is no more that this value on average.
- The RBG method, which can be seen as a straightforward improvement of the real-world operation, requires almost negligible computing times to generate feasible solutions. However, it creates considerable gaps relative to the solution values of our method, increasing with the scale of the instances. Specifically, the robotic workload can on average be reduced by 35.86% (over one-third), which can translate into significant cost savings in terms of the number of robots required or operating expenses of distribution centres, while also boosting the order picking efficiency.
- For DBI, while it does optimize the solution value derived from RBG to some extent, it is dwarfed by our method. The ability of DBI almost reaches only half of ours. In addition, the advantage of DBI over our method in terms of computing time shrinks as the size of the instance increases. In other words, for larger-scaled cases in practice, the reasonable order allocation mechanism within our SA framework achieves further efficiency improvement without significantly increasing the computational time.

**6.3 Sensitivity analysis**

We further generate some managerial insights and operating suggestions from the following sensitivity analysis, where the computation results are derived from our method SA.

6.3.1 Effect of workbench capacity

In this section we try to investigate the impact of workbench capacity on picking efficiency using the total required rack visits as a measure. For the range of values of $C \in [1, 15]$, we conduct ten experiments for each value of $C$ and use the average result as a reference for comparison, while the other warehousing parameters are kept constant as follows: $n = 1000, m = 5, |R| = 500,$ and $\beta = 20$.

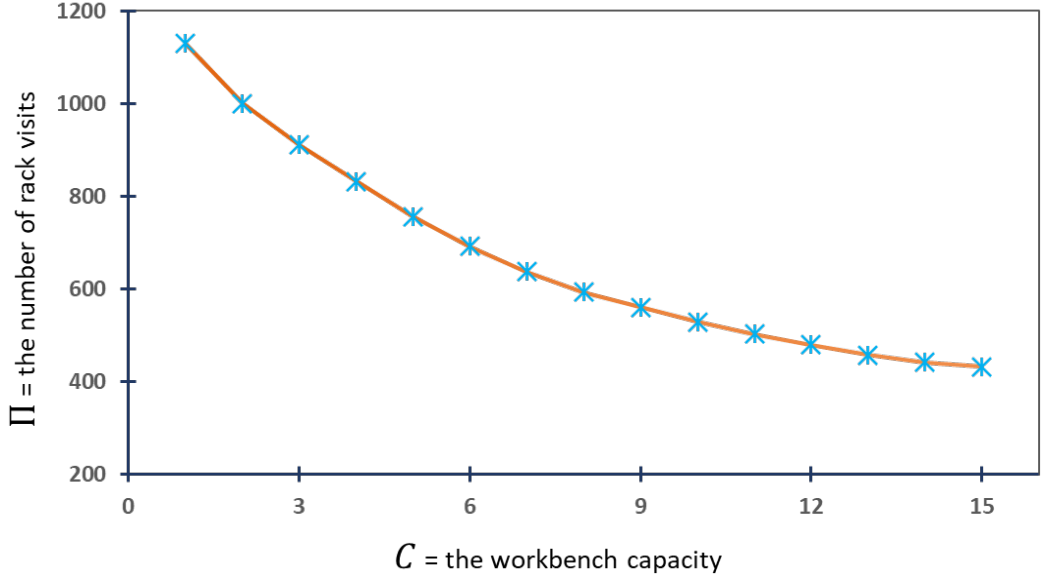

**Figure 6.** Effect of workbench capacity $C$.

Figure 6 shows that as the workbench capacity increases, the number of required rack visits drops significantly, especially when the capacity is relatively small. However, the benefit of investing in additional capacity to further reduce rack movements will start to diminish. This phenomenon can be approximately explained by "a little flexibility goes a long way" (Jordan & Graves, 1995). It also shows that there is an inverse relationship between the intensity of robotic labour and human labour. Therefore, to optimize the efficiency of the picking system reasonably, it is crucial for the manager to strike a balance between these two types of labour.

6.3.2 Effect of picking task quantity

We then explore the impact of the number of assigned orders on picking efficiency, where the evaluation measure is designed by the ratio of the number of orders and the number of rack visits, i.e., the order fulfillments during one time rack movement. We set the values as $n = 250, 500, 750, \dots 2500$, i.e., the number of orders processed by each workstation ranges from 50 to 500, and for each value, we conduct ten experiments and use the average result as a reference for comparison. The other parameters are given and kept constant as follows: $m = 5, C = 15, |R| = 500,$ and $\beta = 20$.

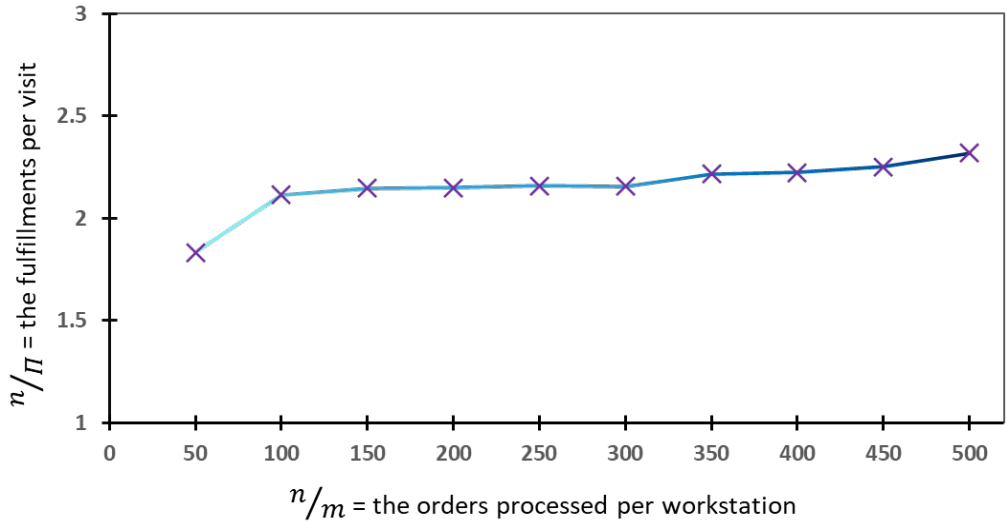

**Figure 7.** Effect of order quantity processed per workstation.

We note from Figure 7 that when each workstation handles a larger volume of orders, the order fulfillment ratio becomes greater, resulting in improved picking efficiency due to the benefit of pooling, i.e., larger volumes of orders can take advantage of economies of scale. However, it is worth noting that this benefit comes at a price, such as longer waiting times for orders to arrive. If the picking system can be approximated as a flow system, i.e., order pooling does not require waiting that usually occurs in warehouses with fast turnover rates, then order coordination should be possibly maximized to fully utilize this pooling effect. However, if the order arrival rate is lower than the service rate of the picking system, waiting times should be factored in when utilizing this pooling effect. It is also important to mention that this advantage is limited by the workbench capacity.

## 7 Conclusions

We consider the joint optimization of order assignment and picking station scheduling in the robot-assisted KIVA warehouse. Contrary to the picker-to-parts warehouse, the KIVA system handles the order picking process with static pickers and movable racks delivered to stations by mobile robots. Therefore, the order assignment policy has a direct impact on the subsequent rack selection, whereas picking station scheduling deals with synchronization of the processing sequencing of the assigned orders and arriving racks. These two interrelated decisions collectively determine the total number of robotic tasks, the closely associated relationships between which can facilitate exploitation of either of their synergy.

We formulate the resulting decision problem as an integrated mixed-integer programming model and ascertain its computational complexity. The commercial solver Gurobi can solve small-sized cases and provide a performance benchmark; however, it cannot address large-sized cases in a reasonable time frame. Therefore, we propose a heuristic algorithm, adapting the basic logic of SA containing a BS mechanism, which is effective in tackling large-sized instances with thousands of orders and racks.

We conduct a series of numerical studies to test our developed method and find that it can produce high-quality solutions for small-scale cases. To further evaluate its performance under large-scale scenarios, we generate instance data according to simulations that closely approximate real-world operations. We find that our method can significantly reduce rack visits by over one-third compared with the greedy heuristics. Our computational results can always be achieved within a reasonable time frame, even for large-scale instances, which is particularly valuable for e-commerce warehouses that face increasing workloads.

We further conduct sensitivity analyses of the important configuration parameters, providing valuable management insights for warehouse managers to improve operations. Our findings suggest that expanding workbench capacity can reduce rack visits, albeit with diminishing benefit. In addition, we discover that accumulating more orders at each single workstation can better utilize their synergistic effect, but this needs to be balanced against the negative impact of waiting times on overall efficiency, which requires careful consideration based on the specific operational context of the warehouse. Lastly, it is worth highlighting that order picking is an interconnected operation, and any changes to configuration must be made with a comprehensive understanding of the potential impacts of various factors.

Regarding the robot-assisted KIVA warehouse, future research should focus more on the stochastic nature of the problem. In addition, more holistic problem settings should be tested in future research, where OAPSSP is coupled with rack storage assignment and/or robotic task allocation and traffic planning. The findings of such research will shed light on how best to organize the complicated operations in real-world parts-to-picker warehouses.

**Acknowledgments**

This research was supported by the National Natural Science Foundation of China (NSFC) under grant number 71831001, the Fundamental Research Funds for the Central Universities under grant number 2021YJS073, and Beijing Logistics Informatics Research Base.

**References**

Aarts, H., Verplanken, B., & Van Knippenberg, A. (1997) Habit and information use in travel mode choices. *Acta Psychologica*, 96(1-2), 1-14.

Ardjmand, E., Shakeri, H., Singh, M., & Bajgiran, O.S. (2018) Minimizing order picking makespan with multiple pickers in a wave picking warehouse. *International Journal of Production Economics*, 206, 169-183.

Azadeh, K., De Koster, R., & Roy, D. (2017) Robotized warehouse systems: Developments and research opportunities. *ERIM Report Series Research in Management*, Erasmus Research

Institute of Management (ERS-2017-009-LIS).

Banker, S. (2016) Robots in the warehouse: It's not just Amazon. *Forbes*.

Bartholdi, J.J. III, Hackman, S.T. (2014) Warehouse and distribution science. Release 0.96. Supply Chain and Logistics Institute, https://www.warehouse-science.com/book/.

Batt, R.J., & Gallino, S. (2019) Finding a needle in a haystack: The effects of searching and learning on pick-worker performance. *Management Science*, 65(6), 2624-2645.

Battini, D., Calzavara, M., Persona, A., & Sgarbossa, F. (2017) Additional effort estimation due to ergonomic conditions in order picking systems. *International Journal of Production Research*, 55(10), 2764-2774.

Blum, C. (2005) Beam-ACO—Hybridizing ant colony optimization with beam search: An application to open shop scheduling. *Computers & Operations Research*, 32(6), 1565-1591.

Boysen, N., & Zenker, M. (2013) A decomposition approach for the car resequencing problem with selectivity banks. *Computers & Operations Research*, 40(1), 98-108.

Boysen, N., Briskorn, D., & Emde, S. (2017) Parts-to-picker based order processing in a rack-moving mobile robots environment. *European Journal of Operational Research*, 262(2), 550-562.

Boysen, N., De Koster, R., & Weidinger, F. (2019) Warehousing in the e-commerce era: A survey. *European Journal of Operational Research*, 277(2), 396-411.

Černý, V. (1985) Thermodynamical approach to the traveling salesman problem: An efficient simulation algorithm. *Journal of Optimization Theory and Applications*, 45, 41-51.

Çeven, E., & Gue, K.R. (2017) Optimal wave release times for order fulfillment systems with deadlines. *Transportation Science*, 51(1), 52-66.

Cho, H.S., Paik, C.H., Yoon, H.M., & Kim, H.G. (2005) A robust design of simulated annealing approach for mixed-model sequencing. *Computers & Industrial Engineering*, 48(4), 753-764.

CNN Business (2018) Life inside an Amazon fulfillment center. Available from https://www.youtube.com/watch?v=iXxPabWb9nI last accessed July 12, 2020.

D'Andrea, R., & Wurman, P. (2008) Future challenges of coordinating hundreds of autonomous vehicles in distribution facilities. *Proceedings of 2008 IEEE International Conference on Technologies for Practical Robot Applications* (pp. 80-83), IEEE.

De Koster, R., Le-Duc, T., & Roodbergen, K.J. (2007) Design and control of warehouse order picking: A literature review. *European Journal of Operational Research*, 182(2), 481-501.

Enright, J.J., & Wurman, P.R. (2011) Optimization and coordinated autonomy in mobile fulfillment systems. *Proceedings of the 25th AAAI Conference on Artificial Intelligence*.

Garey, M.R., & Johnson, D.S. (1979) *Computers and Intractability*. San Francisco: Freeman.

Gu, J., Goetschalckx, M., & McGinnis, L.F. (2010) Research on warehouse design and performance evaluation: A comprehensive review. *European Journal of Operational Research*, 203(3), 539-549.

Jordan, W.C., & Graves, S.C. (1995) Principles on the benefits of manufacturing process flexibility. *Management Science*, 41(4), 577-594.

Kirkpatrick, S., Gelatt Jr, C.D., & Vecchi, M.P. (1983) Optimization by simulated annealing. *Science*, 220(4598), 671-680.

Kumawat, G.L., & Roy, D. (2021) A new solution approach for multi-stage semi-open queuing networks: An application in shuttle-based compact storage systems. *Computers & Operations Research*, 125, 105086.

Lamballais, T., Roy, D., & De Koster, M.B.M. (2017) Estimating performance in a robotic mobile fulfillment system. *European Journal of Operational Research*, 256(3), 976-990.

Masson, R., Lehuédé, F., & Péton, O. (2013) An adaptive large neighborhood search for the pickup and delivery problem with transfers. *Transportation Science*, 47(3), 344-355.

Merschformann, M., Xie, L., & Li, H. (2017) RAWSim-O: A simulation framework for robotic mobile fulfillment systems. *arXiv preprint arXiv:1710.04726*.

Mokotoff, E. (2001). Parallel machine scheduling problems: A survey. *Asia-Pacific Journal of Operational Research*, 18(2), 193.

Neuhub (2019) From: https://neuhub.jd.com/innovation/type/AGV last accessed July 12, 2021.

Ow, P.S., & Morton, T.E. (1988) Filtered beam search in scheduling. *International Journal of Production Research*, 26(1), 35-62.

Pan, J.C.H., Shih, P.H., & Wu, M.H. (2015) Order batching in a pick-and-pass warehousing system with group genetic algorithm. *Omega*, 57, 238-248.

Qin, H., Xiao, J., Ge, D., Xin, L., Gao, J., He, S., ... & Carlsson, J. G. (2022) JD.com: Operations research algorithms drive intelligent warehouse robots to work. *INFORMS Journal on Applied Analytics*, 52(1), 42-55.

Roodbergen, K.J., & Vis, I.F. (2009) A survey of literature on automated storage and retrieval systems. *European Journal of Operational Research*, 194(2), 343-362.

Ropke, S., & Pisinger, D. (2006) An adaptive large neighborhood search heuristic for the pickup and delivery problem with time windows. *Transportation Science*, 40(4), 455-472.

Sabuncuoğlu, İ., Gocgun, Y., & Erel, E. (2008) Backtracking and exchange of information: Methods to enhance a beam search algorithm for assembly line scheduling. *European Journal of Operational Research*, 186(3), 915-930.

Scholz, A., Schubert, D., & Wäscher, G. (2017) Order picking with multiple pickers and due dates–Simultaneous solution of order batching, batch assignment and sequencing, and picker routing problems. *European Journal of Operational Research*, 263(2), 461-478.

Shen, M., Tang, C.S., Wu, D., Yuan, R., & Zhou, W. (2020) JD.com: Transaction-level data for the 2020 MSOM data driven research challenge. *Manufacturing & Service Operations Management*.

Sina, From: https://finance.sina.com.cn/chanjing/gsnews/2020-11-12/doc-iiznezxs1358721.shtml last accessed July 7, 2021.

Valle, C.A., & Beasley, J.E. (2021) Order allocation, rack allocation and rack sequencing for pickers in a mobile rack environment. *Computers & Operations Research*, 125, 105090.

Van Gils, T., Ramaekers, K., Caris, A., & De Koster, R.B. (2018) Designing efficient order picking systems by combining planning problems: State-of-the-art classification and review. *European Journal of Operational Research*, 267(1), 1-15.

Weidinger, F. (2018) A precious mess: On the scattered storage assignment problem. *Operations Research Proceedings 2016* (pp. 31-36). Springer, Cham.

Weidinger, F., & Boysen, N. (2018) Scattered storage: How to distribute stock keeping units all around a mixed-shelves warehouse. *Transportation Science*, 52(6), 1412-1427.

Weidinger, F., Boysen, N., & Briskorn, D. (2018) Storage assignment with rack-moving mobile robots in KIVA warehouses. *Transportation Science*, 52(6), 1479-1495.

Winkelhaus, S., Grosse, E.H., & Morana, S. (2021) Towards a conceptualisation of Order Picking 4.0. *Computers & Industrial Engineering*, 159, 107511.

Wulfraat, M. (2012) Is the Kiva system a good fit for your distribution center? An unbiased distribution consultant evaluation. *MWPVL International White Paper*.

Wurman, P.R., D'Andrea, R., & Mountz, M. (2008) Coordinating hundreds of cooperative, autonomous vehicles in warehouses. *AI Magazine*, 29(1), 9-9.

Xie, L., Thieme, N., Krenzler, R., & Li, H. (2021) Introducing split orders and optimizing operational policies in robotic mobile fulfillment systems. *European Journal of Operational Research*, 288(1), 80-97.

Yang, X., Hua, G., Hu, L., Cheng, T.C.E., & Huang, A. (2021) Joint optimization of order sequencing and rack scheduling in the robotic mobile fulfilment system. *Computers & Operations Research*, 135, 105467.

Yuan, R., Graves, S.C., & Cezik, T. (2019) Velocity-based storage assignment in semi-automated storage systems. *Production and Operations Management*, 28(2), 354-373.

Zaerpour, N., Yu, Y., & De Koster, R. (2017) Small is beautiful: A framework for evaluating and optimizing live-cube compact storage systems. *Transportation Science*, 51(1), 34-51.

Zhuang, Y., Zhou, Y., Yuan, Y., Hu, X., & Hassini, E. (2021) Order picking optimization with rack-moving mobile robots and multiple Workstations. *European Journal of Operational Research*.